\documentclass[conference]{IEEEtran}
\IEEEoverridecommandlockouts

\usepackage[T1]{fontenc}
\usepackage{cite}
\usepackage{amsmath,amssymb,amsfonts,mathtools}
\usepackage{bm}
\usepackage{graphicx}
\usepackage{xcolor}
\usepackage[caption=false,font=footnotesize,labelfont=rm,textfont=rm]{subfig}
\usepackage{booktabs}
\usepackage{url}
\usepackage[bookmarks=false]{hyperref}
\IfFileExists{orcidlink.sty}{%
  \usepackage{orcidlink}%
}{%
  \newcommand{\orcidlink}[1]{}%
}

\hypersetup{
  hidelinks,
  pdftitle={Physics-Aware Conditional SetGAN for Spatially Consistent Multi-User TR 38.901 Channel Generation},
  pdfauthor={Mauro Gonzalo Tarazona-Levano and David Lopez-Perez and Nicola Piovesan and David Gomez-Barquero},
  pdfsubject={IEEE conference paper},
  pdfkeywords={Generative AI, channel modeling, digital twins, 3GPP TR 38.901, Sionna, generative adversarial networks, set transformers, spatial consistency, shadow fading}
}

\graphicspath{{figures/plots/}}

\newcommand{\E}{\mathbb{E}}

\newcommand{\abs}[1]{\left\lvert #1 \right\rvert}

\newcommand{\vecop}{\mathrm{vec}}

\newcommand{\figurefileplaceholder}[1]{%
  \fbox{%
    \parbox[c][3.2cm][c]{0.30\textwidth}{%
      \centering
      \footnotesize Expected vector panel\\[0.35em]
      \texttt{\detokenize{#1}}%
    }%
  }%
}
\newcommand{\autopanelpdf}[1]{%
  \IfFileExists{figures/plots/#1.pdf}{%
    \includegraphics[width=0.31\textwidth]{#1.pdf}%
  }{%
    \figurefileplaceholder{#1.pdf}%
  }%
}

\title{Physics-Aware Conditional SetGAN for Spatially Consistent Multi-User TR 38.901 Channel Generation}

\title{Physics-Aware Conditional SetGAN for Spatially Consistent Multi-User TR 38.901 Channel Generation}

\author{
\IEEEauthorblockN{
Mauro~Gonzalo~Tarazona-Levano\IEEEauthorrefmark{1},
David~Lopez-Perez\IEEEauthorrefmark{1}\IEEEauthorrefmark{2},
Nicola~Piovesan\IEEEauthorrefmark{3}, and
David~Gomez-Barquero\IEEEauthorrefmark{1}
}
\normalsize\IEEEauthorblockA{\emph{
\IEEEauthorrefmark{1}Institute of Telecommunications and Multimedia Applications (iTEAM), Universitat Polit\`ecnica de Val\`encia (UPV), Spain}
}
\normalsize\IEEEauthorblockA{\emph{
\IEEEauthorrefmark{2}Beihang Valencia Polytechnic Institute (BVPI), China}
}
\normalsize\IEEEauthorblockA{\emph{
\IEEEauthorrefmark{3}Huawei Technologies, France}
}

\textit{mgtarlev@iteam.upv.es}

\thanks{This research is supported by the Generalitat Valenciana through the CIDEGENT PlaGenT, Grant CIDEXG/2022/17, Project iTENTE, and the action CNS2023-144333, financed by MCIN/AEI/10.13039/501100011033 and the European Union ``NextGenerationEU''/PRTR.}
}

\begin{document}
\maketitle

\begin{abstract}
TR~38.901-based channel models such as Sionna are reliable, but generating many multi-user channel realizations remains expensive. This paper asks a practical question: can a trained generative model produce multi-user TR~38.901 channels faster than Sionna without losing the spatial correlations imposed by user geometry? To answer this question, we propose a physics-aware, geometry-conditioned SetGAN trained on Sionna reference data. The method separates large-scale received power from normalized small-scale fading, compresses the latter with principal component analysis, and learns the conditional channel distribution in a latent space while preserving geometry-dependent correlations. On the UMa/NLoS benchmark, the model keeps the received-power distributions close to the reference, with about $0.41$~dB Wasserstein distance, and reproduces spatial-consistency profiles with mean deviations below $0.03$ on median curves versus distance. In addition, it reduces elapsed generation time by a factor of $3.45$ and CPU-total cost by a factor of $6.15$ relative to Sionna under matched user positions in the fixed-position CPU-vs-CPU benchmark. These results show that a trained generative model can substantially accelerate TR~38.901 channel generation without breaking the spatial consistency needed to evaluate multi-user systems.
\end{abstract}


\section{Introduction}
\label{sec:intro}
Generating multi-user multiple-input multiple-output (MIMO) channel realizations is computationally expensive, especially when large numbers of user-equipment (UE) deployments and configurations must be evaluated. At the same time, these realizations cannot be treated as independent channel draws: nearby UE should exhibit correlated large-scale effects and coherent small-scale fading structure, since spatial consistency is essential in multi-user MIMO evaluations. This requirement is particularly important when studying multi-user behavior under realistic geometry.

Tools implementing the stochastic channel models specified in 3GPP TR~38.901 provide this type of physically grounded channel generation~\cite{3gpp38901}. Among them, Sionna is an open-source wireless simulation library that implements these models and is widely used as a practical reference~\cite{sionna}. Later studies have also examined the spatial-consistency behavior and calibration of these models~\cite{eval_spatial_consistency,flexible_3gpp}. In that sense, TR~38.901-based simulation tools provide a reliable reference for generating spatially consistent multi-user channels. The practical problem is that repeatedly calling such tools to generate very large numbers of realizations can become prohibitively expensive in runtime.

The central question of this paper is therefore the following: can a generative model be designed to reproduce the multi-user MIMO channel frequency responses generated by a TR~38.901-compliant reference tool, while generating them faster than the simulator itself? In this work, Sionna is used as that reference implementation. In the multi-user setting, this is not only about reproducing realistic coefficients for each UE separately. The real challenge is to reproduce the joint geometry-conditioned structure of a channel realization, so that nearby UE preserve the spatial correlations imposed by the underlying channel model~\cite{deepsets,settransformer}.

\textbf{Related Work and Positioning.}
Generative wireless channel modeling has received growing attention in recent years. In the case of generative adversarial networks (GANs), prior work has covered several related but narrower objectives, including autonomous channel-distribution learning from measurements~\cite{gan_wireless_mag}, conditional prediction of covariance-related quantities~\cite{gan_cov}, reproduction of impulse-response statistics in MIMO channels~\cite{mimo_gan}, and GAN-based modeling from measured massive-MIMO data as an alternative to stochastic models or ray tracers~\cite{gan_massive_mimo}. However, these approaches typically focus on individual links, derived objects such as covariances, or global statistics learned from measurements, rather than on full geometry-conditioned multi-user realizations with explicit spatial consistency across nearby UE.

In parallel, diffusion models have become another strong generative baseline in wireless communications because of their stable training and conditional-generation capabilities~\cite{diffusion_wireless,diffusion_wireless_sota}. Recent examples include position-conditioned statistical CSI generation within a \emph{Digital Twin of Channel} framework~\cite{dtoc} and position-conditioned synthesis of high-dimensional wireless channels for downstream tasks such as compression and beam alignment~\cite{gen_highdim_diffusion}. However, these approaches remain primarily UE-specific, statistical, or task-oriented, rather than targeting full geometry-conditioned TR~38.901 multi-user snapshots with explicit spatial-consistency profiles and direct simulator-cost comparison.

From an architectural viewpoint, the problem requires a suitable inductive bias: a multi-user snapshot is a set of UE, so its representation should be invariant or equivariant to permutations~\cite{deepsets}. Because interactions among UE also matter, a Set-Transformer-style architecture is natural for modeling them expressively as a function of relative geometry~\cite{settransformer}. Likewise, positional encoding through random Fourier features provides a justified way to represent spatial positions by approximating shift-invariant kernels with low-dimensional random maps~\cite{rahimi_recht}.

From the optimization perspective, training generative models over structured, high-dimensional distributions remains delicate. Wasserstein GAN and spectral normalization provide a standard basis for stabilizing adversarial training and controlling the discriminator~\cite{wgan,sn}. Beyond that global stability, works such as Pareto GAN and Top-$k$ training show that rare or extreme regions of the distribution require explicit treatment~\cite{pareto,topk}. This is especially relevant here because received-power tail fidelity and spatial-profile matching depend precisely on regularizing regions of the generative space that the standard adversarial term does not always cover well.

Taken together, the literature shows that generative models and suitable tools already exist for structured wireless data. What is still missing, and what this paper resolves, is a surrogate that combines geometry-conditioned generation of full TR~38.901 multi-user snapshots, explicit preservation of spatial consistency through distance-profile matching, and direct fidelity/runtime benchmarking against a matched-position Sionna reference; this combination is the novelty claimed here.

\begin{figure*}[t]
  \centering
  \includegraphics[width=\textwidth]{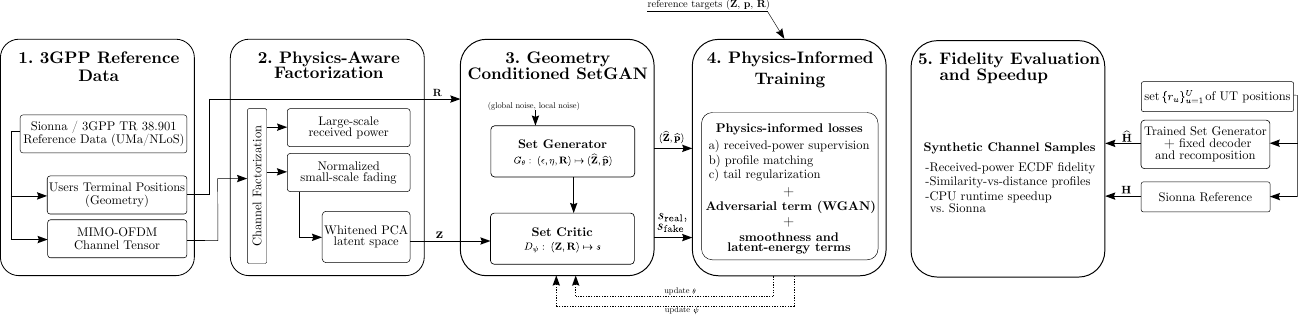}
  \caption{End-to-end workflow of the proposed UMa/NLoS channel surrogate: Sionna reference generation, physics-aware channel factorization, geometry-conditioned SetGAN training, and fidelity evaluation against Sionna under matched UE positions.}
  \label{fig:pipeline_overview}
\end{figure*}

\textbf{Pipeline Overview and Scope.}
Fig.~\ref{fig:pipeline_overview} summarizes the end-to-end workflow in five stages: 1) generation of TR~38.901 reference data with Sionna; 2) physics-aware factorization of the channel into large-scale received power and normalized small-scale fading; 3) geometry-conditioned SetGAN learning in a whitened principal-component latent space; 4) physics-informed training with adversarial, profile-matching, and tail-aware objectives; and 5) fidelity and runtime validation against the reference simulator. Stages 1 and 2 are formalized in Section~\ref{sec:dataset}, stages 3 and 4 are developed in Section~\ref{sec:model}, and stage 5 is addressed through the experimental protocol in Section~\ref{sec:eval} and the final results in Section~\ref{sec:results}.

\section{System Model and Data Representation}
\label{sec:dataset}

This section defines the learning object and the data representation used by the model. The target is a multi-user frequency-domain MIMO-OFDM snapshot consistent with 3GPP TR~38.901, generated in this work through Sionna's implementation of that specification. Starting from these raw channel realizations, the representation used for learning applies two operations. First, each per-UE channel is decomposed into a large-scale received-power term and a normalized small-scale fading tensor. Second, only the normalized fading tensor is compressed in a whitened principal-component space, while the large-scale term is kept as a direct prediction target. The goal of this representation is to align the learning problem with the physical structure of the channel: the slowly varying received-power term is modeled separately, while dimensionality reduction is applied only to the normalized fading component.

\subsection{Reference Snapshot Generation with Sionna}
We generate reference channels with Sionna's implementation of 3GPP TR~38.901 for the deployment scenario considered in this study~\cite{3gpp38901,sionna}. Each channel realization contains one serving BS and an unordered set of $U$ user equipment (UE) with positions $\{\bm r_u\}_{u=1}^{U}$, where $\bm r_u\in\mathbb{R}^{d_r}$. Pathloss and shadow fading remain enabled, so the resulting channel responses already contain the large-scale effects prescribed by the TR~38.901 model.

After removing the singleton batch dimension, the raw frequency-response snapshot is stored as a tensor
\begin{equation}
\mathcal{H}\in\mathbb{C}^{U\times N_{\mathrm{rx,ant}}\times N_{\mathrm{tx}}\times N_{\mathrm{tx,ant}}\times T\times F},
\end{equation}
where $U$ is the number of UE in the snapshot, $N_{\mathrm{rx,ant}}$ is the number of receive antenna branches per UE, $N_{\mathrm{tx}}$ is the number of transmitting entities included in the realization, $N_{\mathrm{tx,ant}}$ is the number of transmit antenna branches per transmitter, $T$ is the number of OFDM symbols, and $F$ is the number of subcarriers.

For learning, the antenna-related axes are merged into a single branch axis. The effective channel of UE $u$ is therefore represented as
\begin{equation}
\bm H_u \in \mathbb{C}^{N_a\times T\times F}.
\end{equation}
Here,
\begin{equation}
N_a \;=\; N_{\mathrm{rx,ant}}\,N_{\mathrm{tx}}\,N_{\mathrm{tx,ant}},
\end{equation}
so that all per-UE antenna branches remain available to the model while the multi-user dimension is treated as an unordered set. This reshaping preserves the full per-UE time-frequency structure and only reorganizes how the antenna-related indices are presented to the learning pipeline.

\subsection{Large-Scale and Small-Scale Channel Decomposition}
The first transformation separates the slowly varying attenuation from the fast channel fluctuations by introducing a large-scale term together with a normalized small-scale fading tensor. The large-scale term is obtained from a per-user received-power scalar:
\begin{equation}
p_u^{\mathrm{lin}}
\;\triangleq\;
\frac{1}{N_aTF}\sum_{a=1}^{N_a}\sum_{t=1}^{T}\sum_{f=1}^{F}\abs{\bm H_u(a,t,f)}^2,
\label{eq:pl_lin}
\end{equation}
and its dB representation
\begin{equation}
p_u^{\mathrm{dB}} \;=\; 10\log_{10}\!\left(p_u^{\mathrm{lin}}\right).
\label{eq:pl_db}
\end{equation}
The normalized small-scale fading tensor is then defined as
\begin{equation}
\widetilde{\bm H}_u \;=\; \frac{\bm H_u}{\sqrt{p_u^{\mathrm{lin}}}},
\label{eq:ssf_norm}
\end{equation}
and the normalized large-scale term is defined as
\begin{equation}
p_u \;\triangleq\; \frac{p_u^{\mathrm{dB}}-\mu_p}{\sigma_p}.
\label{eq:pl_norm}
\end{equation}
This yields a scalar large-scale target and a normalized fading tensor for the subsequent principal-component stage.

\subsection{Small-Scale Fading Compression in Principal Component Space}
\label{sec:pca}
Once the normalized small-scale fading tensor has been isolated, it remains too high-dimensional to model directly. This is the main reason for introducing principal component analysis (PCA): even after removing the large-scale term, direct adversarial learning in the original ambient dimension would be unnecessarily difficult and inefficient. Principal component analysis is therefore fit only on normalized fading vectors, not on the full channel with the large-scale term included. Following that same choice, we split the complex coefficients into $N_c$ real-valued channels, vectorize the normalized small-scale fading tensor, and fit an incremental PCA on a calibration set of $N_{\mathrm{fit}}$ channel realizations:
\begin{equation}
\bm x_u \;=\;
\begin{bmatrix}
\Re\{\vecop(\widetilde{\bm H}_u)\}\\
\Im\{\vecop(\widetilde{\bm H}_u)\}
\end{bmatrix}
\in \mathbb{R}^{N_x},
\qquad
N_x=N_cN_aTF,
\label{eq:vec}
\end{equation}
We retain $K$ components and whiten the PCA coefficients,
\begin{equation}
\bm c_u = \bm W^\top(\bm x_u-\bm\mu)\in\mathbb{R}^{K},
\qquad
\bm z_u = \bm c_u \odot \bm\lambda^{-\frac12}\in\mathbb{R}^{K}.
\label{eq:pca_whiten}
\end{equation}
The generator operates in this whitened latent space, while a fixed decoder reconstructs the normalized small-scale fading tensor:
\begin{equation}
\widehat{\bm x}_u = \bm W\left(\widehat{\bm z}_u \odot \bm\lambda^{\frac12}\right) + \bm\mu.
\label{eq:pca_decode}
\end{equation}
Keeping the decoder fixed isolates the learnable part to the conditional generator and makes linear PCA reconstruction a useful reference in the results.

\section{Proposed Physics-Aware Conditional Set-Based Generator}
\label{sec:model}

Once the target representation is fixed, the next step is to learn a surrogate that maps user-terminal geometry to synthetic channel realizations in that latent space. The design goal is not only to generate plausible samples, but to do so in a way that preserves the spatial-consistency structure of the Sionna reference. In the implementation used here, this is enforced through both the architecture and the loss: the generator uses geometry-aware set attention with a relative positional bias, and training explicitly matches distance-dependent similarity profiles computed over UE pairs. The subsections below describe the conditional set formulation, the geometry-aware generator, and the training objective that differentiates the model from a generic GAN.

\subsection{Problem Formulation}
\label{subsec:problem}
Each channel realization is treated as an unordered set of $U$ user terminals with geometry
$\mathbf{R}=[\bm r_1,\ldots,\bm r_U]^\top\in\mathbb{R}^{U\times d_r}$.
For each user terminal $u$, the target variables are a whitened PCA latent
$\bm z_u\in\mathbb{R}^{K}$ representing the normalized small-scale fading tensor and a scalar
$p_u\in\mathbb{R}$ representing the normalized large-scale term.
Collecting the per-user latents gives
\begin{equation}
\mathbf{Z}=[\bm z_1,\ldots,\bm z_U]^\top \in \mathbb{R}^{U\times K}.
\end{equation}
The generator maps global noise $\bm\epsilon$, local noises $\{\bm\eta_u\}_{u=1}^{U}$,
and geometry $\mathbf{R}$ to synthetic per-user outputs:
\begin{equation}
\begin{aligned}
(\widehat{\mathbf{Z}},\widehat{\bm p})
&= G_\theta(\bm\epsilon,\{\bm\eta_u\}_{u=1}^{U},\mathbf{R}), \\
\widehat{\mathbf{Z}} &\in \mathbb{R}^{U\times K}, \qquad
\widehat{\bm p}\in\mathbb{R}^{U}.
\end{aligned}
\label{eq:setgan_generator}
\end{equation}
Because the user-terminal ordering is arbitrary, the model is designed to be permutation-consistent: re-indexing the input set induces the same re-indexing on the outputs.

\subsection{Geometry-Aware Set Generator}
\label{subsec:gen}
The generator combines random-Fourier positional encoding of the user-terminal coordinates, shared global and per-user local noise injection, and a set-attention backbone with geometry-dependent interactions. In the implementation, the coordinate embedding is passed through a learned MLP, combined with realization-level and user-level noise, and then processed by a spatial set encoder. The initial per-user hidden state is
\begin{equation}
\bm h_u^{(0)} = \bm e_u + \alpha\,g(\bm\epsilon) + \ell(\bm\eta_u),
\qquad u=1,\ldots,U,
\label{eq:gen_in}
\end{equation}
where $\bm e_u$ is the learned embedding of $\bm r_u$, $g(\bm\epsilon)$ is a realization-level perturbation broadcast to all user terminals, and $\ell(\bm\eta_u)$ captures user-specific randomness.

Context is built through multi-head self-attention. For head $h$, the interaction logit between user terminals $i$ and $j$ is
\begin{equation}
L_{ij}^{(h)}=
\frac{(\bm q_i^{(h)})^\top \bm k_j^{(h)}}{\sqrt{d_k}}
+ b(\bm r_i,\bm r_j),
\label{eq:att_block}
\end{equation}
where $b(\bm r_i,\bm r_j)$ is a shared pairwise bias. This additive term imposes a distance-aware interaction prior, favoring stronger coupling between nearby user terminals. The attention mechanism is not purely content-based: its logits include a learned relative positional bias that depends on pairwise user geometry. This bias combines an explicit distance-decay term with a nonlinear correction learned from relative coordinates, so that nearby user terminals are encouraged to interact more strongly while still allowing departures from a purely monotone distance law. Two output heads then predict $\widehat{\bm z}_u$ and $\widehat{p}_u$ for each user terminal. A fixed PCA decoder reconstructs $\widehat{\widetilde{\bm H}}_u$ from $\widehat{\bm z}_u$, and the predicted power term is re-applied to form the full channel sample when needed:
\begin{equation}
\widehat{\bm H}_u \;=\; \widehat{\gamma}_u\,\widehat{\widetilde{\bm H}}_u,
\qquad
\widehat{\gamma}_u \triangleq 10^{(\sigma_p\widehat p_u+\mu_p)/20}.
\label{eq:recompose}
\end{equation}

\subsection{Set Critic and Learning Objective}
\label{sec:loss}
The critic $D_\psi(\mathbf{Z},\mathbf{R})$ is a permutation-invariant set encoder followed by mean pooling and an MLP head. Spectral normalization is applied to stabilize Wasserstein training. With real latents $\mathbf{Z}$ and generated latents $\widehat{\mathbf{Z}}$, the critic loss is
\begin{equation}
\mathcal{L}_D =
\E\!\left[D_\psi(\widehat{\mathbf{Z}},\mathbf{R})\right]
-\E\!\left[D_\psi(\mathbf{Z},\mathbf{R})\right]
+\lambda_{\mathrm{drift}}\,\E\!\left[D_\psi(\mathbf{Z},\mathbf{R})^2\right].
\label{eq:LD}
\end{equation}
The corresponding adversarial generator term is
\begin{equation}
\mathcal{L}_{\mathrm{adv}} = -\E\!\left[D_\psi(\widehat{\mathbf{Z}},\mathbf{R})\right].
\label{eq:Ladv}
\end{equation}

To target the deployment metrics of interest, the adversarial term is augmented with physics-aware auxiliaries that are central to the proposed method. The key mechanism used to preserve spatial consistency is not only the set-based architecture, but the explicit matching of geometry-dependent user-pair statistics during training. Concretely, from the decoded fading tensors and predicted received-power scalars, we build pairwise similarity matrices over all user pairs and bin them by user separation distance. The losses $\mathcal{L}_{\mathrm{ff}}$ and $\mathcal{L}_{\mathrm{pl}}$ then force the generated snapshots to match the reference fading-structure and received-power similarity profiles versus distance. In addition, $\mathcal{L}_{\mathrm{smooth}}$ penalizes abrupt local power variations among nearby users, $\mathcal{L}_{\mathrm{sup}}$ regresses the large-scale received-power term directly, $\mathcal{L}_{\mathrm{topk}}$ and $\mathcal{L}_{\mathrm{mass}}$ regularize the received-power tail, and $\mathcal{L}_{\mathrm{en}}$ controls the energy of the whitened latent space. The full generator objective is
\begin{equation}
\begin{aligned}
\mathcal{L}_G
&= \mathcal{L}_{\mathrm{adv}}
+ \lambda_{\mathrm{sup}}\mathcal{L}_{\mathrm{sup}}
+ \lambda_{\mathrm{smooth}}\mathcal{L}_{\mathrm{smooth}}
+ \lambda_{\mathrm{ff}}\mathcal{L}_{\mathrm{ff}}
+ \lambda_{\mathrm{pl}}\mathcal{L}_{\mathrm{pl}} \\
&\quad
+ \lambda_{\mathrm{topk}}\mathcal{L}_{\mathrm{topk}}
+ \lambda_{\mathrm{mass}}\mathcal{L}_{\mathrm{mass}}
+ \lambda_{\mathrm{en}}\mathcal{L}_{\mathrm{en}}.
\end{aligned}
\label{eq:LG}
\end{equation}
This objective is the main mechanism by which the model is pushed to preserve geometry-dependent spatial consistency: rather than matching only marginal channel statistics, it also aligns pairwise similarity structure as a function of user separation distance.

\section{Experimental Setup and Evaluation Protocol}
\label{sec:eval}

Unless stated otherwise, the reported results correspond to the tail-aware training path in Section~\ref{sec:model}, i.e., the generator objective in \eqref{eq:LG} with profile matching and the tail-sensitive terms $\mathcal{L}_{\mathrm{topk}}$ and $\mathcal{L}_{\mathrm{mass}}$.

For the benchmark considered here, the main protocol variables are $U$ (UE per snapshot), $N_{\mathrm{fit}}$ (PCA fitting set size), $N_{\mathrm{train}}$ (GAN training set size), $K$ (latent dimension), $B_{\mathrm{mb}}$ (mini-batch size), $N_{\mathrm{eval}}$ (unpaired evaluation set size), $e^\star$ (selected checkpoint epoch), and $N_{\mathrm{rep}}$ (runtime repeats). Checkpoint selection follows a coarse-to-fine automatic sweep over training epochs. Candidate epochs are first filtered by received-power distribution metrics, and the final checkpoint is selected from the acceptable candidates using a normalized joint score over those metrics and the two spatial-profile curve errors. The baseline is Sionna itself, and training/evaluation geometries use the same in-distribution random-square sampler; other learned generators and deliberately out-of-distribution layouts such as hotspots are outside the present scope.

We report three metric families. First, received-power empirical cumulative distribution functions (ECDFs) are compared through the Kolmogorov-Smirnov (KS) statistic, Wasserstein-1 distance in dB, and quantile errors summarized as MAE$_Q$. Second, spatial consistency is assessed through distance-binned median/IQR profiles over all $U(U-1)/2$ UE pairs within each channel realization, using both a fading-structure similarity profile and a received-power similarity profile. For each unordered pair $(i,j)$, we compute the separation distance $d_{ij}=\lVert \bm r_i-\bm r_j\rVert_2$ and a similarity score $s_{ij}^{(m)}$, where $m\in\{\mathrm{ff},\mathrm{pl}\}$ denotes fading-structure or received-power similarity. For the power profile, $s_{ij}^{(\mathrm{pl})}=\exp(-(\Delta p_{ij}^{\mathrm{dB}})^2/(2\sigma_{\mathrm{sf}}^2))$, where $\Delta p_{ij}^{\mathrm{dB}}$ is the received-power difference in dB and $\sigma_{\mathrm{sf}}$ is the shadow-fading standard deviation; for the fading profile, $s_{ij}^{(\mathrm{ff})}$ is the normalized inner product between vectorized normalized receive-correlation matrices. For distance bins $\{\mathcal{B}_b\}_{b=1}^{B}$, the profile summaries are
\begin{equation}
\begin{aligned}
\tilde{s}_b^{(m)}
&=\operatorname{median}\!\left\{s_{ij}^{(m)}: d_{ij}\in\mathcal{B}_b\right\},\\
\operatorname{IQR}_b^{(m)}
&=\operatorname{IQR}\!\left\{s_{ij}^{(m)}: d_{ij}\in\mathcal{B}_b\right\}.
\end{aligned}
\label{eq:profile_bin_summary}
\end{equation}
The discrepancy between generated and reference median profiles is then summarized by
\begin{equation}
\begin{aligned}
\mathrm{MAE}_{\mathrm{curve}}^{(m)}
&=\frac{1}{B}\sum_{b=1}^{B}\left|\tilde{s}_{b,\mathrm{GAN}}^{(m)}-\tilde{s}_{b,\mathrm{REAL}}^{(m)}\right|,\\
\Delta_{\max}^{(m)}
&=\max_b\left|\tilde{s}_{b,\mathrm{GAN}}^{(m)}-\tilde{s}_{b,\mathrm{REAL}}^{(m)}\right|.
\end{aligned}
\label{eq:profile_curve_metrics}
\end{equation}
For both metrics, a good match preserves the reference median/IQR profile across distance bins. For received-power similarity, this includes non-monotonic features such as a dip followed by a later rise; for fading-structure similarity, it means reproducing the reference decay and spread with distance. A poor match would flatten, shift, or over/under-spread either profile. Third, runtime is measured with a fixed-position batched benchmark using $N_{\mathrm{rep}}$ repeats in CPU-vs-CPU and GPU-vs-GPU modes, so simulator and surrogate are compared under identical geometry.

\section{Results and Discussion}
\label{sec:results}

For the UMa/NLoS benchmark considered here, the results answer the paper's central question affirmatively: the learned surrogate remains close to the Sionna reference on the target fidelity metrics, supporting repeated Monte Carlo evaluation at lower generation cost. Although the pipeline can be instantiated for other TR~38.901 scenarios and the set-based architecture does not fix $U$, the present validation uses a controlled UMa/NLoS setting with $U=100$ user terminals, $K=1024$ PCA coefficients, $N_{\mathrm{eval}}=750$ snapshots, random-square side length $10$--$2000$~m, and selected checkpoint $e^\star=60$; cross-scenario and variable-UE evaluations are left outside the current scope. The subsections below detail the received-power, spatial-consistency, and fixed-position runtime evidence behind that claim.

\subsection{Received-Power Distribution}
Received-power fidelity for UMa/NLoS is summarized by $\mathrm{KS}=0.0170$, $W_1=0.406$~dB, and MAE$_Q=0.431$~dB over $75{,}000$ per-user-terminal power samples per domain. The quantile errors show a remaining lower-tail offset at the $5$th percentile, while the central and upper quantiles stay within about $0.42$~dB.

\begin{figure}[t]
\centering
\includegraphics[width=\columnwidth]{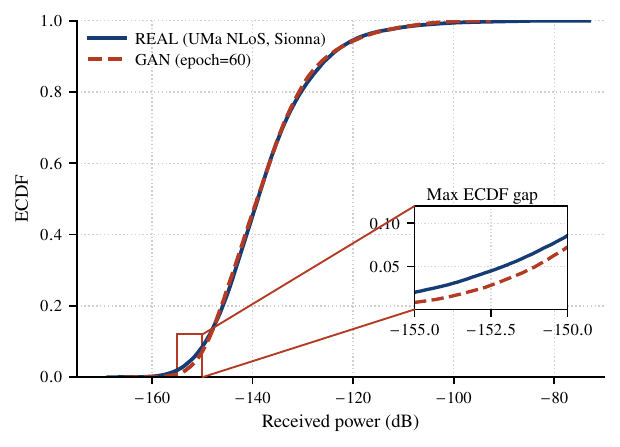}
\caption{Received-power ECDF for the UMa/NLoS benchmark. The curves compare the Sionna reference and the GAN at the selected checkpoint $e^\star=60$; the inset highlights the maximum ECDF-gap region.}
\label{fig:ecdf_uma}
\end{figure}

This distribution-level agreement is visible in Fig.~\ref{fig:ecdf_uma}: the Sionna and GAN curves nearly overlap, while the inset localizes the maximum ECDF gap.

\subsection{Spatial-Consistency Profiles}
\begin{figure}[t]
\centering
\subfloat[Fading-structure similarity versus user-terminal separation.\label{fig:profiles_uma_fading}]{%
\includegraphics[width=\columnwidth]{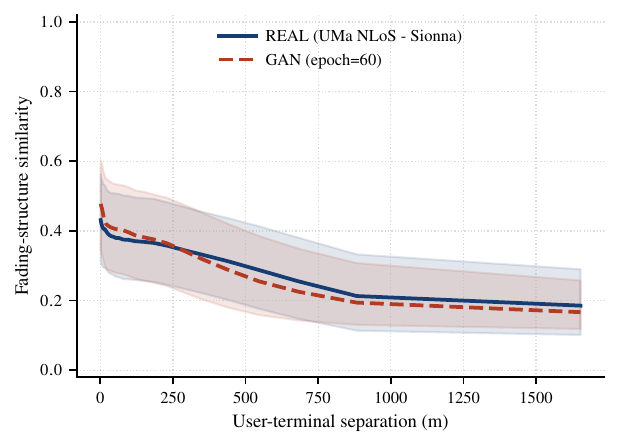}%
}\\[0.4em]
\subfloat[Received-power similarity versus user-terminal separation.\label{fig:profiles_uma_power}]{%
\includegraphics[width=\columnwidth]{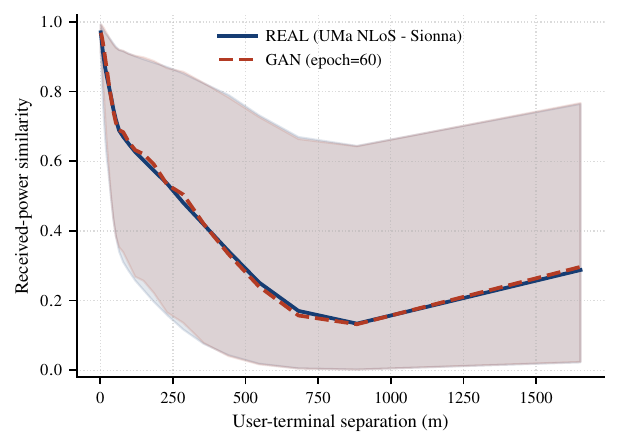}%
}
\caption{Spatial-consistency profiles for the UMa/NLoS benchmark. In both panels, lines show the median profile and shaded bands the interquartile range; the curves compare the Sionna reference and the GAN at $e^\star=60$.}
\label{fig:profiles_uma}
\end{figure}

Fig.~\ref{fig:profiles_uma} reports the two distance-binned spatial-consistency profiles defined in \eqref{eq:profile_bin_summary} and summarized by \eqref{eq:profile_curve_metrics}. For the received-power similarity profile, the median-curve MAE is $0.0255$ with worst-bin deviation $0.0518$. For the fading-structure similarity profile, the median-curve MAE is $0.0260$ with worst-bin deviation $0.0496$. Because both similarity metrics are normalized to the $[0,1]$ range, these errors are small in absolute terms: the curve-MAEs correspond to deviations of only 2.5 and 2.6 percentage points, and even the worst-bin deviations remain at 5.2 and 5.0 percentage points. Both metrics are computed from $3{,}712{,}500$ user-terminal pairs per domain, which makes the bin-level trends stable. The GAN follows the same distance-dependent trends as Sionna, with moderate degradation at larger separations.

The PCA dimension is therefore chosen from a fidelity--runtime tradeoff, as summarized in Table~\ref{tab:k_runtime_uma}.

\subsection{Runtime Benchmark}
All fixed-position runtime benchmarks were run on an Intel Core i9-14900K workstation with an NVIDIA RTX A6000 (49~GB VRAM, driver 580.126.09). Here, generation time denotes elapsed wall-clock time of channel generation only, whereas call time includes full benchmark overhead. With matched user positions and $100$ repeats, CPU-vs-CPU yields the largest relative gain: $3.45\times$ in generation time, $6.15\times$ in CPU-total cost, and $1.88\times$ in call time. Under the same protocol in GPU-vs-GPU mode, with both processes capped at $16$~GB, the surrogate still achieves $2.52\times$, $1.31\times$, and $1.22\times$ speedups, respectively, although the relative gain is smaller because the Sionna reference also benefits from GPU acceleration.

\begin{table}[t]
\caption{UMa/NLoS fidelity--runtime tradeoff in CPU-vs-CPU mode.}
\label{tab:k_runtime_uma}
\centering
\scriptsize
\setlength{\tabcolsep}{5pt}
\begin{tabular}{cccc}
\toprule
$K$ & $W_1$ [dB] & $\mathrm{MAE}_{\mathrm{curve}}^{(\mathrm{ff})}$ & \shortstack{Gen\\speedup} \\
\midrule
256  & 0.660 & 0.0542 & $3.87\times$ \\
512  & 0.436 & 0.0435 & $3.61\times$ \\
1024 & 0.406 & 0.0260 & $3.45\times$ \\
\bottomrule
\end{tabular}
\end{table}

\section{Conclusion}
\label{sec:concl}
For the UMa/NLoS benchmark considered here, the answer to the paper's central question is yes: a geometry-conditioned SetGAN can reproduce the main multi-user TR~38.901/Sionna statistics closely enough to be useful while generating samples at lower cost than direct simulator calls. The selected model preserves received-power and spatial-consistency metrics while reaching $3.45\times$ generation-time and $6.15\times$ CPU-total speedups in CPU-vs-CPU mode, and $2.52\times$ and $1.31\times$ under GPU-vs-GPU execution with a $16$~GB memory cap per process. These conclusions are limited to the matched-position UMa/NLoS protocol studied here; broader claims across other TR~38.901 scenarios, PCA dimensions, or deployment settings require additional validation.

\bibliographystyle{IEEEtran}
\bibliography{bib/references}

\end{document}